\documentclass[10pt,twocolumn,letterpaper]{article}

\usepackage{cvpr}              % To produce the CAMERA-READY version
\definecolor{cvprblue}{rgb}{0.21,0.49,0.74}

\usepackage[pagebackref,breaklinks,colorlinks,allcolors=cvprblue]{hyperref}
\usepackage{multirow}

\def\paperID{15349} % *** Enter the Paper ID here
\def\confName{CVPR}
\def\confYear{2026}

\title{Med-SORA: Symptom to Organ Reasoning in Abdomen CT Images}

\author{
You-Kyoung Na\\
Chonnam National University\\
{\tt\small youkyoung@jnu.ac.kr}
\and
Yeong-Jun Cho\\
Chonnam National University\\
{\tt\small yj.cho@jnu.ac.kr}
}

\begin{document}
\maketitle
\begin{abstract}
Understanding symptom-image associations is crucial for clinical reasoning. 
However, existing medical multimodal models often rely on simple one-to-one hard labeling, oversimplifying clinical reality where symptoms relate to multiple organs.
In addition, they mainly use single-slice 2D features without incorporating 3D information, limiting their ability to capture full anatomical context.
In this study, we propose Med-SORA, a framework for symptom-to-organ reasoning in abdominal CT images. 
Med-SORA introduces RAG-based dataset construction, soft labeling with learnable organ anchors to capture one-to-many symptom–organ relationships, and a 2D–3D cross-attention architecture to fuse local and global image features. 
To our knowledge, this is the first work to address symptom-to-organ reasoning in medical multimodal learning.
Experimental results show that Med-SORA outperforms existing medical multimodal models and enables accurate 3D clinical reasoning. The code and dataset are available at \url{https://WILL_BE_SOON}
\end{abstract}

\section{Introduction}
\label{sec:intro}
% 이 연구란
Electronic Medical Records~(EMRs) are structured clinical documents that include patient symptoms, diagnoses, treatments, and other healthcare-related information, as recorded by healthcare professionals. Understanding the connection between patient symptoms in EMRs and medical images is an important research goal for advancing our understanding of clinical decision-making. Such connections can also serve as supporting evidence for diagnosis, helping AI systems make more accurate and interpretable decisions.
However, linking symptom descriptions in text to findings in medical images is technically challenging due to the gap between abstract language and visual representations.
\looseness=-1

\begin{figure} [t]
    \includegraphics[width=\linewidth]{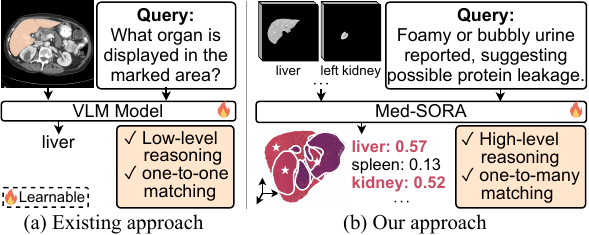}
    \caption{Comparison between existing medical vision-language models and the proposed Med-SORA. Ground-truth organs are marked with $\star$.}
    \label{fig:fig1_compare}
\vspace{-15pt}
\end{figure}

Recent studies have explored medical multimodal models in various applications, including pathology image analysis~\cite{yu2024cp}, medical visual question answering (Med-VQA)~\cite{hu2024omnimedvqa}, medical report generation~\cite{liu2021auto}, and interactive diagnostic systems~\cite{gao2024training}.
However, existing research has several limitations. 
First, as shown in Fig.~\ref{fig:fig1_compare}(a), current medical multimodal models perform low-level reasoning by identifying organs in marked image areas using one-to-one text-image correspondence. 
Such approaches overlook the one-to-many nature of symptom–organ associations, oversimplifying clinical reality.
For example, a symptom like `jaundice' can relate to multiple organs (e.g., liver, gallbladder, pancreas). 
Second, existing studies~\cite{eslami2023pubmedclip, zhang2023biomedclip} rely mainly on single 2D slice-based features and lack joint 2D-3D analysis, limiting their ability to capture full anatomical context and spatial relationships.

% ---------- fig 1-----------

% 본 연구의 제안 기법 - 1. 데이터셋, 2. soft labeling, 3. 3d-2d cross attn
To address these limitations, we propose the Symptom-to-Organ Reasoning (Med-SORA), which performs high-level clinical reasoning by inferring related organs from patient symptom descriptions as shown in Fig.~\ref{fig:fig1_compare}(b).
Unlike existing approaches that rely on direct visual pattern matching, our method reflects clinical reality where symptoms can probabilistically relate to multiple organs.
To this end, we first construct a symptom–text dataset by extracting symptom–organ knowledge from reliable medical literature using Retrieval-Augmented Generation (RAG). 
Second, we propose soft labeling with learnable organ anchors to model probabilistic relationships between symptoms and multiple organs. 
Third, we fuse 2D and 3D image features from abdominal CT images to capture both fine-grained slice-level detail and global 3D anatomical structure.
Finally, Med-SORA is trained to align symptom-text and image features in a shared embedding space.

% 실험적 성과 - 1. 데이터셋, 2. soft labeling, 3. 3d-2d cross attn 순서
To validate our method, we conducted experiments using a RAG-based symptom text dataset (validated by medical experts) and the publicly available BTCV abdominal CT dataset~\cite{btcv2015miccai}.
Results show that soft labeling more effectively captures complex organ relationships in the embedding space than hard labeling.
Moreover, our 2D–3D cross-attention architecture outperformed single-modality 2D and 3D models in organ identification accuracy.
Med-SORA achieved the highest overall performance compared to existing medical multimodal models.
Unlike prior studies that focus on low-level tasks such as image classification or organ detection, Med-SORA performs high-level clinical reasoning.
It directly infers and visualizes symptom-related organs in 3D space, enabling intuitive spatial interpretation.

% 주요 기여
The main contributions of this work are as follows:
\begin{itemize}
    \item We propose Med-SORA, a framework that aligns symptom text with 3D CT images for clinical reasoning.
    \item We build a reliable and clinically valid symptom-organ dataset using RAG from trusted medical literature.
    \item We introduce soft labeling with learnable organ anchors to model one-to-many symptom–organ relationships.
    \item We design a 2D-3D cross-attention architecture to extract image features from CT scans for better understanding of abdominal organs.
    \item Med-SORA outperforms existing multimodal medical models on the symptom-to-organ reasoning task.
\end{itemize}
To the best of our knowledge, Med-SORA is the first attempt to perform symptom-to-organ reasoning in the medical multimodal domain.

\section{Related Works}
\label{sec:related_work}

\subsection{RAG-based Construction of Medical Datasets}
Retrieval Augmented Generation (RAG) integrates external information retrieval to overcome the limitations of pre-trained language models that rely solely on their internal knowledge~\cite{lewis2020retrieval}. 
This approach enables generative models to search and reference relevant external documents, thereby enhancing factual accuracy and currency of the generated content.
The medical field especially benefits from RAG because it can access the latest medical literature, keeping up with fast-changing research and treatment guidelines~\cite{jin2023medcpt}.
As a result, RAG-based approaches~\cite{zakka2024almanac, wu2025graph} have been actively explored in medical applications, including clinical question answering and medical knowledge retrieval.
Studies~\cite{jin2019pubmedqa, krithara2023bioasq} have also employed retrieval-based approaches to automatically construct datasets from large-scale repositories such as PubMed and Wikipedia.

While these approaches show potential for automated dataset construction, they primarily focus on generating data for factual knowledge retrieval and simple question-answering, making them inadequate for constructing datasets requiring complex clinical reasoning.

\subsection{Multimodal Learning in the Medical Domain}
CLIP~\cite{radford2021clip}, a representative vision-language model~(VLM), learns semantic associations between images and text by training on large-scale image–text pairs.
CLIP’s effectiveness in learning generalizable image–text representations has led to the development of domain-specific variants for medical applications.
For example, PubMedCLIP~\cite{eslami2023pubmedclip} adapts CLIP to the medical domain by retraining it on PubMed literature and medical images, and BioMedCLIP~\cite{zhang2023biomedclip} further extends this approach using broader biomedical data to support diverse medical tasks.
% (yjcho) soft labeling에 대한 내용을 related works에 어떻게 작성할지 고민 필요.
%Ko et al.~\cite{ko2025bringing} addressed semantic overlap in medical data by combining clinically-enhanced dynamic soft labeling with negation-based hard negatives.
These medical VLMs have been applied to various downstream tasks, including medical visual question answering~\cite{xu2025structure}, diagnostic imaging assistance~\cite{nori2025sequential}, and medical education~\cite{shaw2025artificial}.

Reasoning across multiple modalities (i.e., text and image) is a key challenge, especially in the medical field where accurate decision-making depends on integrating various types of information. In clinical settings, reasoning involves combining symptoms, imaging results, and medical knowledge to reach a diagnosis.
Large language models like GPT~\cite{radford2018GPT} have advanced reasoning capabilities, driving progress in multimodal reasoning when combined with image understanding.
Recently, text-based reasoning models have also been developed specifically for the medical domain.
% Large language models like GPT~\cite{radford2018GPT} have advanced reasoning capabilities and accelerated multimodal reasoning research when combined with image understanding.
MedMCQA~\cite{pal2022medmcqa} provides a benchmark dataset for evaluating medical reasoning through multiple-choice clinical questions. 
Building on such datasets, LLM-based models like Med-PaLM~\cite{singhal2025medpalm} demonstrate strong performance by generating answers through reasoning grounded in medical knowledge. 
Huy et al.~\cite{huy2025interactive} similarly proposed a concept-based similarity reasoning network for medical image interpretation, which learns region-specific patterns and models spatial interactions.

However, existing medical reasoning models have significant limitations when applied to real clinical scenarios. Text-based models like Med-PaLM~\cite{singhal2025medpalm} excel at medical knowledge reasoning but cannot process visual information from medical images, limiting their applicability in image-dependent diagnoses. Conversely, medical vision-language models focus primarily on basic image understanding tasks rather than complex clinical reasoning that integrates symptoms with imaging findings.

% ------------------- fig ---------------------
\begin{figure*} [t]
    % \centering
    % \hspace*{0.15\textwidth}% <-- 왼쪽에 15%의 여백 추가
    \includegraphics[width=\textwidth]{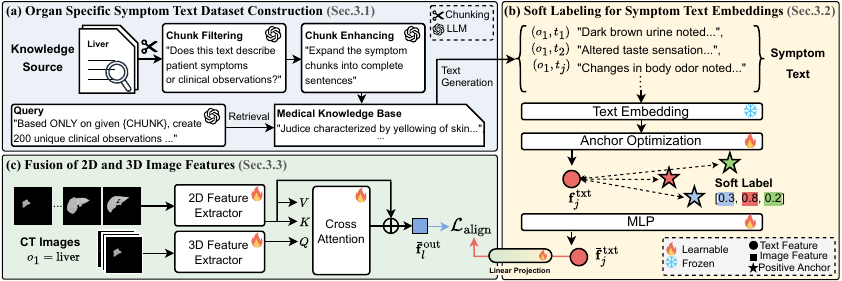}
    \caption{{The pipeline of Med-SORA, showing the optimization process for $o_1=\text{liver}$. It consists of (a) RAG-based dataset construction, (b) soft labeling-based text embedding learning, and (c) 3D-2D feature-based image embedding learning with text-image embedding alignment.}}
    \label{fig:proposed_framework}
\vspace{-15pt}
\end{figure*}
% ------------------- fig ---------------------

\section{Proposed Methods}
\label{sec:proposed_methods}
We propose a multimodal framework, Med-SORA, that performs Symptom-to-Organ Reasoning in abdominal CT images. 
Med-SORA takes patient symptom text as input and infers and visualizes related organ locations in 3D abdominal CT images, as demonstrated in Fig.~\ref{fig:fig1_compare}(b).
The proposed Med-SORA consists of three main components: 
(a) organ-specific symptom text dataset construction, 
(b) soft labeling for symptom text via learnable organ anchors, and 
(c) fusion of 2D and 3D image features.
The overall pipeline of the proposed method is shown in Fig.~\ref{fig:proposed_framework}.

\subsection{Organ-Specific Symptom Text Dataset \\ Construction} 
\label{sec:text_dataset}

To enable multimodal learning between symptom text and abdominal images, a high-quality text dataset of organ-specific symptoms is required.
However, such datasets are limited due to the high cost and need for medical expertise.
To address this, we construct a symptom text dataset using Retrieval-Augmented Generation (RAG)~\cite{lewis2020retrieval}, as shown in Fig.~\ref{fig:proposed_framework}(a).

Our symptom text dataset covers $N$ major abdominal organs, $\mathcal{O} = \{o_1, o_2, \ldots, o_{N}\}$, including the liver, kidney, and stomach.
For each organ, symptom texts are collected from PubMed~\cite{pubmed} and Wikipedia~\cite{wikipedia}.
PubMed provides specialized, clinically validated medical information, whereas Wikipedia offers more general descriptions.
By leveraging both sources, we aim to capture a wide spectrum of symptom texts.
Both platforms support keyword-based retrieval through their official APIs.
Accordingly, we used organ-specific keywords to extract relevant symptom texts for each target organ.
To standardize sentence length and support efficient processing, the raw symptom texts are divided into chunks of two to three sentences. 

Some of the retrieved texts after chunking may include sentences that are not related to symptoms.
To improve quality, a Large Language Model (LLM) is used to either rewrite them into clinically relevant symptom descriptions or remove them if they contain irrelevant information or lack meaningful clinical content.
To achieve this, we prompt the LLM with two instructions:
(1) \textit{``Does this text describe a patient’s symptom or clinical observation?''} and
(2) \textit{``Based ONLY on given \{\texttt{CHUNK}\}, create 200 unique clinical observations described in complete, medical-style sentences.''}
These prompts guide the LLM to refine and 
restructure the texts to better reflect relevant medical information.
As a result, we generate $N^{\text{txt}}$ symptom texts, $\mathbf{T}_i = \{t_{i1}, t_{i2}, \cdots, t_{iN^{\text{txt}}}\}$, for each organ $o_i$, without explicitly mentioning the organ name, enabling inference based solely on symptom information.
The quality and reliability of the generated symptom text dataset were validated by medical professionals.
The detailed data construction process is described in the supplementary materials.

\subsection{Soft Labeling for Symptom Text Embeddings}
\label{sec:soft_labeling}
A symptom text is often related to multiple organs rather than just one.
For example, feeling uncomfortable after eating is intuitively associated with the stomach, but may also involve other organs, such as the pancreas and gallbladder, which contribute to the digestive process.
Therefore, soft labeling is more suitable than hard labeling for learning symptom–organ embedding spaces.
Soft labeling methods~\cite{gao2024softclip, ko2025bringing} have been proposed, but they rely on a simple inter-data similarity or predefined threshold.
Such methods are hard to capture the complex and overlapping relationships between symptoms and organs.

% ------------------- fig ---------------------
\begin{figure}[t]
    \centering
    \hspace{\linewidth}
    \includegraphics[width=\linewidth]{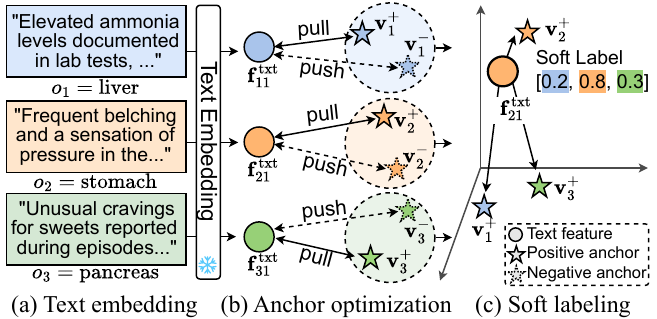}
    \caption{{Soft labeling for symptom text embeddings via learnable anchors.} Different colors represent organ classes. $\leftrightarrow$ means similarity computation, while pull/push operations show the optimization process for $s^+$ and $s^-$ with margin $m$.}
    \label{fig:add_soft_label}
\vspace{-15pt}
\end{figure}
% ------------------- fig ---------------------

To address this limitation, we propose a learnable anchor-based soft labeling to model complex symptom–organ associations as shown in Fig.~\ref{fig:add_soft_label}.
Using a {pretrained text embedding model~\cite{neelakantan2022openai}, the symptom text $t_{ij}$ for each organ is embedded into vectors $\mathbf{f}^\text{txt}_{ij} \in \mathbb{R}^{d_\text{txt}}$ (Fig.~\ref{fig:add_soft_label}(a)). $i$ and $j$ denote the indexes of the organ and the symptom text, respectively.
We then define learnable positive and negative anchors $\mathbf{v}_{i}^+,  \mathbf{v}_{i}^- \in \mathbb{R}^{d_{\text{txt}}}$ for each organ $o_i$ to capture complex symptom–organ relationships.
As shown in Fig.~\ref{fig:add_soft_label}(b), these anchors are optimized during training: $\mathbf{v}_{i}^+$ learns the distinctive symptom representations specific to organ $o_i$, while $\mathbf{v}_{i}^-$ captures general symptom representations shared across other organs.

The positive and negative similarities between the anchors and the symptom embedding $\mathbf{f}^{\text{txt}}_{ij}$ are calculated as follows:
\begin{equation}
    s^+_{i}(\mathbf{f}^\text{txt}_{ij}) = \text{sim}(\mathbf{v}_{i}^+, \mathbf{f}^\text{txt}_{ij}), \quad
    s^-_{i}(\mathbf{f}^\text{txt}_{kj}) = \text{sim}(\mathbf{v}_{i}^-, \mathbf{f}^\text{txt}_{kj}),
\label{eq:1}
\end{equation}
where $i \neq k$, and the function $\text{sim}(\textbf{a}, \textbf{b}) = \frac{\mathbf{a} \cdot \mathbf{b}}{||\mathbf{a}|| \cdot ||\mathbf{b}||}$ represents cosine similarity.
Inspired by triplet loss~\cite{hoffer2015deep}, we define a margin function as follows:
\begin{equation}
\small
    \mathcal{M}\left(s^+_{i}, s^-_{i}\right) = \max \left(0, m-s^+_{i}\right) + \max\left(0, s^-_{i}- (1-m)\right),
\label{eq:2}
\end{equation}
where $m$ denotes the margin of the loss.
This function enforces symptom text embeddings to achieve similarity $s_{i}^+$ with positive anchors above margin $m$ and similarity $s_{i}^-$ with negative anchors below $(1-m)$.
To train the positive and negative anchors $(\mathbf{v}^+_i,\mathbf{v}^-_i)$ of the organ $o_i$, we optimize the following anchor margin loss.
\begin{equation}
    \mathcal{L}_\text{anchor} = \mathbb{E}_{\mathbf{f}^\text{txt}_{ij} \in \mathcal{P}}[\mathcal{M}(s_{i}^+, s_{i}^-)] +
     \mathbb{E}_{\mathbf{f}^\text{txt}_{kj} \in \mathcal{N}}[\mathcal{M}(s_{i}^-, s_{i}^+)],
\label{eq:3}
\end{equation}
where $\mathcal{P}$ denotes the set of positive symptom vectors associated with organ $o_i$, and $\mathcal{N}$ denotes the set of negative symptom vectors from other organs.
Note that these learnable anchors $(\mathbf{v}^+_i, \mathbf{v}^-_i)$ for each organ $o_i$ are pre-trained in an offline stage using only symptom text data, independently of the main image–text alignment training.

After training the anchors, we compute soft labels for each symptom text $t_{kj}$ with respect to each organ $o_i$ based on the corresponding positive anchor $\mathbf{v}_{i}^+$ by
\begin{equation}
    y^{\text{txt}}_{o_i}(t_{kj}) = \{{\text{sim}\left(\mathbf{v}_{i}^+, \mathbf{f}^\text{txt}_{kj}\right) + 1}\} /{2}.
\label{eq:5}
\end{equation}
Considering multiple possible $N$ organ associations, the soft label vector for each symptom text $t_j$ is defined as $\mathbf{y}^{\text{txt}}(t_{kj}) = [y^{\text{txt}}_{o_1}, y^{\text{txt}}_{o_2},\ldots, y^{\text{txt}}_{o_N}]$.
Unlike one-hot labels that assign each symptom to a single organ, our method allows soft associations with multiple organs as shown in Fig.\ref{fig:add_soft_label}(c).
Instead of using softmax, we normalize the cosine similarity score for each soft label $y^{\text{txt}}_{o_i}$ to the $[0, 1]$ range. 
This avoids enforcing a probability distribution and better captures overlapping organ–symptom relationships.

Based on the obtained soft labels $\mathbf{y}^{\text{txt}}$, we further train a Multi-Layer Perceptron (MLP) to extract symptom text features guided by $\mathbf{y}^{\text{txt}}$.
We define the feature through the MLP by $\bar{\mathbf{f}}^{\text{txt}}=\text{MLP}\left(\mathbf{f}^{\text{txt}}\right)$.
The classification head produces $N$ outputs, one for each class, with sigmoid activation applied to each.
The MLP and head are trained by minimizing a cross-entropy loss $\mathcal{L}_\text{txt}$ between the predicted labels $\hat{\mathbf{y}}^{\text{txt}}=\text{Head}(\bar{\mathbf{f}}^{\text{txt}})$ and the soft labels $\mathbf{y}^{\text{txt}}$.

% ------------------- fig ---------------------
\begin{figure}[t]
    \centering
    \includegraphics[width=\linewidth]{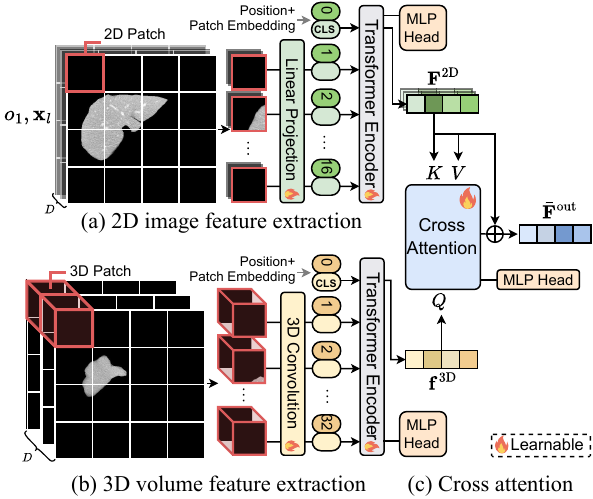}
    \caption{The proposed 2D-3D feature fusion architecture.}
    \label{fig:3d feature extractor}
\vspace{-15pt}
\end{figure}
% ------------------- fig ---------------------

\subsection{Fusion of 2D and 3D Image Features}
\label{sec:image_features}
In this section, we propose a feature extraction framework that jointly captures both 2D and 3D representations from abdominal CT scans.
To enhance organ-level understanding, we extract slice-wise features using a 2D encoder and volume-level context using a 3D encoder, and integrate them through a cross-attention mechanism.

An abdominal CT scan consists of multiple 2D axial slice images, each $l$-th image denoted as $\mathbf{x}_l \in \mathbb{R}^{H \times W}$, where $H$ and $W$ represent the image size.
To focus on major abdominal organs and normalize the varying number of CT slices across patients, we sample $D$ slices for each organ. The resulting input volume of an organ is represented as $\mathbf{X} \in \mathbb{R}^{D \times H \times W}$.
We omit the organ index $i$ for simplicity in this section.

\noindent \textbf{2D image feature extraction.}\quad
To extract 2D features from each slice, we apply a ViT-Large~\cite{dosovitskiy2020vit} model as a 2D feature encoder $E_{\text{2D}}$. This process is illustrated in Fig.~\ref{fig:3d feature extractor}(a).
Inspired by the use of the [CLS] token in ViT for global image representation, we extract the 2D feature of slice $\mathbf{x}_l$ as
\begin{equation}
\mathbf{f}_l^{\text{2D}}= E_{\text{2D}}\left(\left[\mathbf{x}^{\text{cls}},\mathbf{W} \cdot \text{tokenize}(\mathbf{x}_l)\right]+\mathbf{E}^{\text{2D}}_{\text{pos}}\right),
\end{equation}
where $\mathbf{x}^{\text{cls}}$ is a learnable [CLS] token, and each image patch is tokenized and linearly projected using $\mathbf{W}$.
The positional embedding $\mathbf{E}^{\text{2D}}_{\text{pos}}$ is added to the [CLS] token and patch embeddings.
The output $\mathbf{f}_l^{\text{2D}} \in \mathbb{R}^{d_{\text{img}}}$ corresponds to the final embedding of the [CLS] token, which represents the global feature of the slice.
Each image slice $\mathbf{x}_l$ is independently fed into the encoder to extract its 2D feature representation.
As a result, the set of 2D features is given by 
\begin{equation}
\mathbf{F}^{\text{2D}}= \left[\mathbf{f}^{\text{2D}}_1,\mathbf{f}^{\text{2D}}_2,\ldots,\mathbf{f}^{\text{2D}}_D\right] \in \mathbb{R}^{D\times d_{\text{img}}}.    
\end{equation}

\noindent \textbf{3D volume feature extraction.}\quad
For 3D volume feature extraction, we extend the ViT structure to three dimensions, as illustrated in Fig.~\ref{fig:3d feature extractor}(b).
While the 2D ViT processes a single image, our model divides the entire 3D volume into patches of size $(p_D, p_H, p_W)$. To preserve spatial structure, we employ a 3D convolution~\cite{tran20153dcnn} with a 3D kernel for patch embedding, defined as:
\begin{equation}
\mathbf{X}^\text{3D} = \text{Conv3D}\left(\mathbf{X}; \ \text{kernel}=(p_D,p_H,p_W)\right).
\label{eq:3d_patch_embed}
\end{equation}
This operation produces $D/p_D \times H/p_H \times W/p_W$ patches, each embedded into a $d_\text{img}$-dimensional vector space.
To extract the 3D volume representation, we use a transformer-based 3D encoder:
\begin{equation}
    \mathbf{f}^\text{3D} = E_{\text{3D}}\left(\left[\mathbf{x}^{\text{cls}},\mathbf{X}^\text{3D}\right]+\mathbf{E}^{\text{3D}}_\text{pos}\right).
\end{equation}
where $\mathbf{x}^{\text{cls}}$ is a learnable [CLS] token and $\mathbf{E}^{\text{3D}}_\text{pos}$ is a learnable positional embedding.
The 3D encoder is designed to be more lightweight than the 2D encoder, as volumetric data requires higher computational resources.

\noindent \textbf{Cross attention.}\quad
In medical imaging, certain slices often contain critical features of lesions or organs. Therefore, rather than treating all slices equally, it is important to selectively emphasize informative slices based on global context. To this end, we employ cross-attention to integrate slice-wise 2D features with the global 3D context, as illustrated in Fig.~\ref{fig:3d feature extractor}(c). We use the 3D global feature $\mathbf{f}_\text{3D}$ as the query and the 2D features $\mathbf{F}_\text{2D}$ as the keys and values:
\begin{equation}
\mathbf{F}^\text{out} = \text{crossAttn}(Q = \mathbf{f}^\text{3D}, \  K=V= \mathbf{F}^\text{2D}).
\label{eq:cross_attention}
\end{equation}
This allows the model to focus more on the most informative slices.
Finally, we add the features to generate integrated representations by 
\begin{equation}
\bar{\mathbf{F}}^\text{out} = \mathbf{F}^\text{2D} + \mathbf{F}^\text{out} \in \mathbb{R}^{D \times d_\text{img}}.
\label{eq:slice_fusion}
\end{equation}
This fused representation $\bar{\mathbf{F}}^\text{out}$ integrates the original 2D slice features with globally attended feature $\mathbf{F}^\text{out}$, enhancing semantic understanding of both the full scan and fine-grained local details.

\subsection{Loss functions}
For organ classification, we apply separate MLP heads to the 2D slice features $\mathbf{F}^\text{2D}$, 3D volume features $\mathbf{f}^\text{3D}$, and the fused features $\bar{\mathbf{F}}^\text{out}$. Each MLP head produces organ class predictions and is trained using cross-entropy loss between predictions and ground truth labels. The classification losses are defined as $\mathcal{L}_\text{2D}$, $\mathcal{L}_\text{3D}$, and $\mathcal{L}_\text{fusion}$ respectively, and the total image classification loss is:
\begin{equation}
    \mathcal{L}_\text{image} = \mathcal{L}_\text{2D} + \mathcal{L}_\text{3D} + \mathcal{L}_\text{fusion}.
\end{equation}

\noindent \textbf{Image--symptom text alignment loss.} \quad
To align symptom text features $\mathbf{f}^{\text{txt}}$ and image features $\bar{\mathbf{F}}^\text{out}$ in a shared embedding space, we optimize a contrastive loss.
An additional projection layer $\mathbf{W}^{\text{align}} \in \mathbb{R}^{d_\text{img} \times d_\text{txt}}$ is applied to the text branch to match the image feature dimension $d^{\text{img}}$ by $\hat{\mathbf{f}}^\text{txt} = \mathbf{W}^{\text{align}} \cdot \bar{\mathbf{f}}^\text{txt}$.

We define the $l$-th slice image feature of the $i$-th organ by $\bar{\mathbf{f}}^\text{out}_{il} \in \mathbf{\bar{F}}^{\text{out}}_{i}$. 
For effective multimodal learning, we employ InfoNCE~\cite{oord2018infonce} as follows:
\begin{equation}
\small
\mathcal{L}_{\text{align}} = -\sum_{i=1}^{N} \left[ \log \frac{\exp\left(\text{avgSim}(\bar{\mathbf{f}}^{\text{out}}_{il}, \hat{\mathbf{f}}^{\text{txt}}_{ij})/\tau\right)}{\sum_{k=1}^{N} \exp\left(\text{avgSim}(\bar{\mathbf{f}}^{\text{out}}_{il}, \hat{\mathbf{f}}^{\text{txt}}_{kj})/\tau\right)} \right],
\label{eq:infonce_loss}
\end{equation}
where $\tau$ is the temperature scaling factor, and $i\neq k$.
$\text{avgSim}(\cdot)$ denotes an average similarity function defined by
\begin{equation}
    \text{avgSim}(\bar{\mathbf{f}}^\text{out}_{l},\hat{\mathbf{f}}^\text{txt}_{j})=\sum_{l=1}^{D}\sum_{j=1}^{N^{\text{txt}}}\text{sim}(\bar{\mathbf{f}}^\text{out}_{l},\hat{\mathbf{f}}^\text{txt}_{j}),
\end{equation}
where $l$ and $j$ are indexes of the image slice and the symptom text, respectively. $\text{sim}(\cdot)$ is a cosine similarity function.
The final loss is defined as $\mathcal{L}_\text{total} = \mathcal{L}_\text{image}+ \mathcal{L}_\text{align}$, which is optimized to train the image feature extractor and align image and text features in a shared embedding space.

\section{Experimental Results}
\subsection{Datasets and Settings}
\noindent \textbf{Datasets.} \quad For image data, we utilize the BTCV dataset~\cite{btcv2015miccai}, which contains contrast-enhanced abdominal CT images with 11 organ segmentation labels. 
Among them, we used seven main organs: liver, pancreas, kidney, gallbladder, spleen, stomach, and adrenal glands. We excluded non-organ structures such as the esophagus, blood vessels, and veins.
We apply the segmentation masks for seven organs to the original CT images to generate organ-specific image regions for our experiments.
For symptom text data, we use the organ–symptom dataset constructed as described in Sec.~\ref{sec:text_dataset} 
The dataset contains 1,400 symptom descriptions (200 per organ for the selected seven organs) and is split into training and test sets with an 8:2 ratio. The training set uses automatically generated soft labels from Sec.~\ref{sec:soft_labeling}, while the test set is annotated with multi-labels validated by medical experts.

\noindent \textbf{Settings.} \quad All experiments were conducted on a NVIDIA L40S GPU. Training parameters for the text embedding MLP are as follows: epochs – 500, batch size – 32, learning rate – 0.001, optimizer – Adam. The margin value for soft labeling was empirically set to 0.8. 
For the 2D transformer encoder, we used an ImageNet-21k~\cite{ridnik2021imagenet} pre-trained model as backbone and fine-tuned it on the BTCV dataset for 10 epochs. 
For 3D feature extraction, we set patch size $(p_D, p_H, p_W) = (8, 16, 16)$ with six transformer blocks. The 3D transformer and cross-attention module were trained for 50 epochs.
Training parameters for the text-image alignment linear projection layer are as follows: epochs -- 50, InfoNCE loss temperature parameter $\tau$ -- 0.1.
We evaluate performance using Rank-$k$ accuracy ($k=1,2,3$) and mean Average Precision (mAP). 
Rank-$k$ accuracy measures the proportion of cases where the correct answer appears in the top $k$ predictions, while mAP represents the average precision across all organ classes.

\noindent \textbf{Inference.} \quad 
The cosine similarity between the learned positive anchors $\mathbf{v}_{i}^+$ and image features $\bar{\mathbf{F}}^\text{out}$ serves as the inference probability as shown in Fig.~\ref{fig:final_visual}.

\begin{figure}[t]
    \centering
    \includegraphics[width=\linewidth]{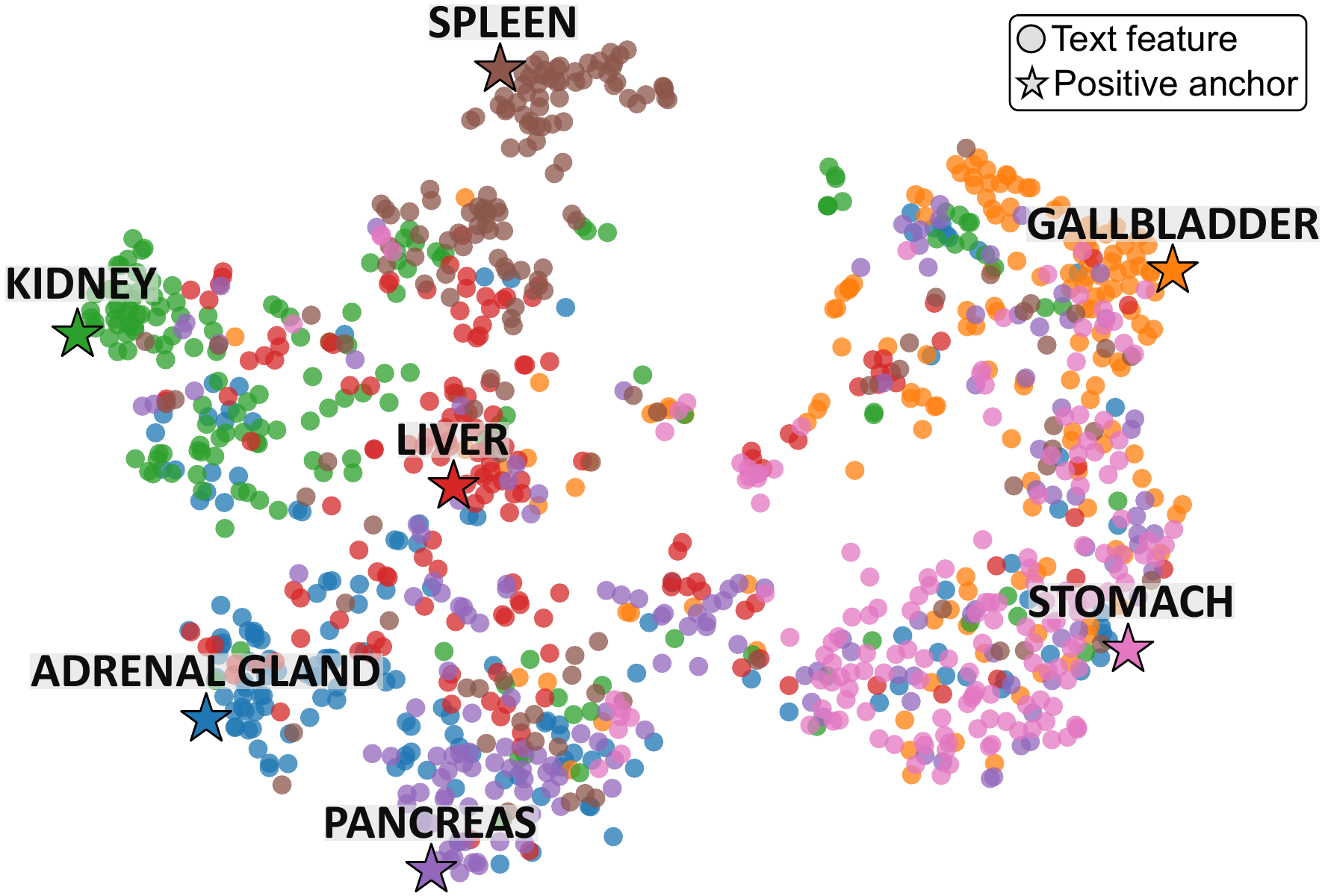}
    \caption{t-SNE visualization of symptom text embeddings learned with soft labeling. Different colors indicate different organ classes, and star symbols denote the learned positive anchors for each organ. (Best viewed in color)}
    \label{fig:soft_embedding}
\end{figure}

\subsection{Effects of the Proposed Methods}

\begin{table}[t]
  \small
  \centering
  \renewcommand{\arraystretch}{0.9} % vertical margin of cell
  \begin{tabular}{c|c|p{5.2cm}}
  \noalign{\hrule height 1pt}
  \textbf{Category} & \textbf{Dist}   & \textbf{Symptom text}   \\ \hline
  \multirow{2}{*}{\begin{tabular}{@{}c@{}}Closest\\Symptom\end{tabular}} &   
  0.74          & Signs of chronic disease of the detoxification organ including palmar erythema and spider angiomas noted. \\ \cline{2-3}
  & 1.55        & Presence of xanthomas on the eyelids reported as a new finding.      \\ \hline
  \multirow{2}{*}{\begin{tabular}{@{}c@{}}Farthest\\Symptom\end{tabular}}
  & 39.41       & Severe abdominal pain in the right upper quadrant noted.   \\ \cline{2-3}
  & 40.31       & Dull abdominal discomfort after eating fatty foods noted by the patient.     \\  \noalign{\hrule height 1pt}
  \end{tabular}
  \caption{Closest and farthest symptom texts from the liver positive anchor $\mathbf{v}^+_1$ in the text embedding space.}
\label{tab:embedding_space_text}
\end{table}

\noindent\textbf{Soft labeling for symptom text embeddings.} \quad 
To verify that complex symptom-organ relationships are effectively learned, we analyze the symptom text embedding space trained with soft labeling. 
Figure~\ref{fig:soft_embedding} shows a t-SNE visualization of text embeddings learned with soft labeling. 
Unlike hard labeling which creates completely separated clusters for each class, soft labeling forms flexible clusters that reflect associations between symptoms and multiple organs. 
Table~\ref{tab:embedding_space_text} shows examples of symptom texts that are closest to and farthest from the liver’s positive anchor in the embedding space, measured by L2 distance.
The closest symptoms consist of liver-specific descriptions directly related to liver function, whereas the farthest symptoms are general abdominal symptoms.
This demonstrates that symptoms strongly associated with a single organ tend to be located near cluster centers, while those related to multiple organs are more likely to appear near cluster boundaries.

\begin{figure}[t]
    \centering
    \includegraphics[width=\linewidth]{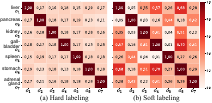}
    \caption{Correlation heatmaps between organs learned using (a) hard labeling and (b) soft labeling.}
    \label{fig:heatmap}
\vspace{-15pt}
\end{figure}

Figure~\ref{fig:heatmap} shows correlation heatmaps between organs learned with hard labeling and soft labeling approaches. 
In hard labeling (Fig.~\ref{fig:heatmap}(a)), each organ is learned independently, resulting in uniformly low correlations between organs. 
In contrast, soft labeling exhibits higher correlations for organ pairs with anatomical proximity or functional relationships, as shown in Fig.~\ref{fig:heatmap}(b).
For example, the heatmap shows that the stomach and liver have strong location-based associations, likely due to their anatomical proximity and shared symptoms such as ``abdominal pain''.
In contrast, the gallbladder and liver show strong functional associations through their common role in bile production.

We evaluate the effectiveness of soft labeling in a text-to-image matching task, where a symptom text is used to retrieve related organs from abdominal CT images.
Table~\ref{tab:soft_label} compares the retrieval performance of hard and soft labeling approaches.
Soft labeling consistently  outperforms hard labeling across all metrics, with larger performance gaps in Rank-2 and Rank-3. 
This suggests that hard labeling focuses on single-organ prediction. 
In contrast, soft labeling better reflects clinical reality, where symptoms may be related to multiple organs, thereby facilitating the learning of multi-organ relationships.

\begin{table}[t]
    \footnotesize
    \centering
    \renewcommand{\arraystretch}{0.9} % vertical margin of cell
    \begin{tabular}{l|r|r|r|r}
    \noalign{\hrule height 1pt}
    \textbf{Labeling}   & \textbf{Rank-1} & \textbf{Rank-2} & \textbf{Rank-3} & \textbf{mAP}   \\ \hline
    Hard      & 74.64           &  82.14         & 86.42           & 67.53              \\
    % \rowcolor{gray!15}
    \textbf{Soft}      & \textbf{77.50}  &  \textbf{89.64} & \textbf{94.85}  &  \textbf{73.46}     \\ \noalign{\hrule height 1pt}
    \end{tabular}
\caption{{Comparison of hard and soft labeling in text-to-image matching performance.}}
\label{tab:soft_label}
\end{table}

\begin{table}[t]
    % \footnotesize
    \small
    \centering
    \renewcommand{\arraystretch}{0.9} % vertical margin of cell
    \begin{tabular}{l|r|r|r|r}
    \noalign{\hrule height 1pt}
    \textbf{Structure}  & \textbf{Rank-1}   & \textbf{Rank-2}   & \textbf{Rank-3}   & \textbf{mAP} \\ \hline
    2D Only             & 75.00             &  86.07            & 94.29             &  71.53       \\
    3D Only             & 33.21             &  60.00            & 74.29             &  39.60        \\
    2D+3D (concat)      & \underline{76.43} & \underline{87.14} & \underline{94.64} &  \underline{72.56}\\
    % \rowcolor{gray!15}
    % \rowcolor{customgray}
    \textbf{2D+3D (cross-Attn)}  & \textbf{77.50}  &  \textbf{89.64}  & \textbf{94.85}  & \textbf{73.46} \\ \noalign{\hrule height 1pt}
    \end{tabular}
\caption{Comparison of different 2D and 3D image feature fusion approaches. The best and second-best scores are marked in bold and {underlined}.}
\vspace{-15pt}
\label{tab:2d/3d comparison}
\end{table}

\noindent\textbf{Fusion of 2D/3D image features.} \quad 
To evaluate the effectiveness of fusing 2D slice-level details with 3D volume-level global information, we compare various image feature fusion methods.
Table~\ref{tab:2d/3d comparison} summarizes the performance comparison of different 2D-3D feature fusion approaches.
Using only 2D features achieves a Rank-1 accuracy of 75\% and an mAP of 86.07\%, while using only 3D features results in significantly lower performance at 33.21\% and 60\%, respectively.
This is because 2D features are extracted independently from each slice, resulting in a larger number of training samples, whereas 3D features are extracted at the volume level only.
Simple concatenation of 2D and 3D features improves performance to 76.43\% Rank-1 accuracy and 72.56\% mAP, compared to using either feature alone.
In contrast, our proposed 2D-3D fusion achieves the best performance, demonstrating more effective integration of fine-grained 2D anatomical details with 3D spatial context.

\subsection{Performance Comparison}

We evaluate the performance of the proposed Med-SORA against existing methods on the symptom text-to-medical image reasoning task, as shown in Tab.~\ref{tab:performance_comparison}.
As no prior work has been explicitly designed for this task, we conduct fair comparisons by adapting representative models from relevant domains. 
These include a general text encoder: BERT~\cite{devlin2019bert}, medical-specific language models: ClinicalBERT~\cite{huang2019clinicalbert}, PubMedBERT~\cite{gu2021pubmedbert}, BioMedGPT~\cite{zhang2024biomedgpt}. 
In addition, we tested multimodal vision-language models (VLM): CLIP~\cite{radford2021clip}, PubMedCLIP~\cite{eslami2023pubmedclip}, BioMedCLIP~\cite{zhang2023biomedclip}.
The experimental settings for adapting these baselines to our task are described in detail in the supplementary materials to ensure fairness.

\begin{table}[t]
    % \small
    % \fontsize{8}{9}\selectfont  % 전체 표의 폰트 크기를 8pt, 줄간격 9pt로 설정
    \small
    \centering
    \renewcommand{\arraystretch}{0.9} % vertical margin of cell
    \begin{tabular}{l|r|r|r|r}
    \noalign{\hrule height 1pt}
    \textbf{Model}              & \textbf{Rank-1}   & \textbf{Rank-2}   & \textbf{Rank-3}   & \textbf{mAP}  \\ \hline
    BERT~\cite{devlin2019bert}          & 60.71             & 81.43            & 87.86             &  51.19        \\
    ClinicalBERT~\cite{huang2019clinicalbert}   & 64.29         & \underline{85.00}        & \underline{91.43}     &  55.58        \\
    PubMedBERT~\cite{gu2021pubmedbert}    & \underline{74.29} & \underline{85.00} & 90.71 & \underline{71.65} \\
    BioMedGPT~\cite{zhang2024biomedgpt}      & 38.21             & 62.50            & 78.92             &  37.40        \\ \hline
    CLIP~\cite{radford2021clip}           & 60.71             &  73.57            & 85.71             & 53.49         \\
    PubMedCLIP*~\cite{eslami2023pubmedclip}   & 47.85             &  60.35            & 83.57             &  38.12        \\ 
    PubMedCLIP~\cite{eslami2023pubmedclip}      & 63.93             &  79.29            & 88.21             &  61.43        \\ 
    BioMedCLIP*~\cite{zhang2023biomedclip}  & 18.93             &  37.50            & 55.36             &  20.02        \\
    BioMedCLIP~\cite{zhang2023biomedclip}  & 23.21             &  40.00            & 54.29             &  23.95        \\ \hline
    \textbf{Ours}   & \textbf{77.50}    &  \textbf{89.64}   & \textbf{94.85}    &  \textbf{73.46}   \\ \noalign{\hrule height 1pt}
    \end{tabular}
    \caption{Performance comparison with existing text and multimodal models on symptom-to-organ reasoning task. * denotes zero-shot performance.}
\vspace{-15pt}
\label{tab:performance_comparison}
\end{table}

Text-based models--BERT, ClinicalBERT, PubMedBERT, and BioMedGPT--were combined with Med-SORA’s image encoder and fine-tuned for the symptom-to-organ matching task.
We evaluated both zero-shot and fine-tuned performance for VLM-based models, especially PubMedCLIP and BioMedCLIP, which have demonstrated strong zero-shot performance in prior studies.
Our method, Med-SORA achieves superior performance across all evaluation metrics compared to other methods.

\begin{figure*}[t!]
    \centering
    \includegraphics[width=\linewidth]{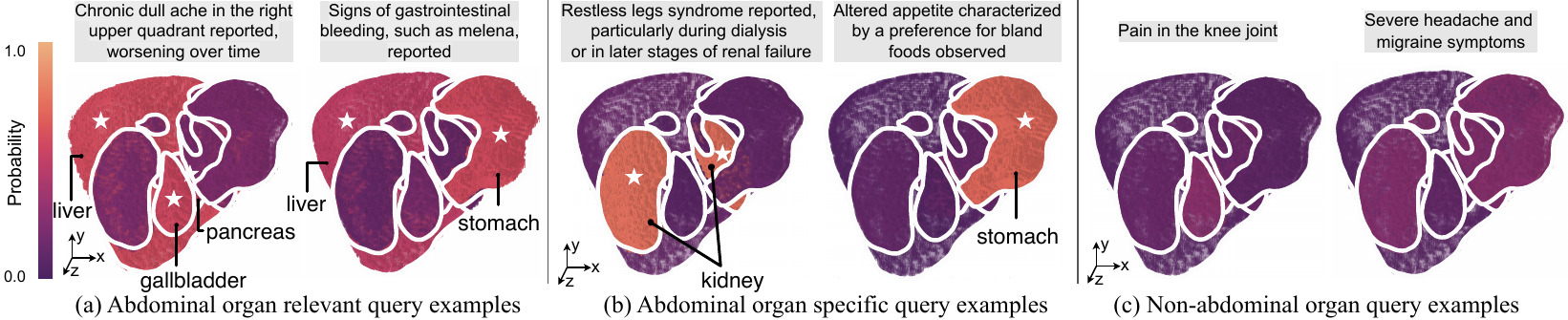}
    \caption{3D probability visualizations generated by Med-SORA for different symptom text queries (shown in gray boxes): (a) symptoms related to multiple organs, (b) organ-specific symptoms, and (c) non-abdominal symptoms. Ground-truth organs are indicated with overlaid star $\star$ markers. Med-SORA effectively handles both single-organ and multi-organ symptom reasoning.}
    \label{fig:final_visual}
\vspace{-15pt}
\end{figure*}

Figure~\ref{fig:final_visual} illustrates the reasoning results of Med-SORA when given symptom text queries.
The model infers the most relevant organ and highlights its 3D segmentation based on the symptom description.
As shown in Fig.~\ref{fig:final_visual}(a), when a symptom query is related to multiple organs, the model assigns probabilities to multiple relevant organs.
In contrast, symptoms that are strongly associated with a specific organ, as in Fig.~\ref{fig:final_visual}(b), result in a distinctly high probability for that organ.
When irrelevant symptom queries such as ``knee joint pain'' or ``headache'' are given as input, as shown in Fig.~\ref{fig:final_visual}(c), all abdominal organs receive uniformly low probability scores.
These results demonstrate that Med-SORA effectively handles both single-organ and multi-organ symptom reasoning, while remaining robust to out-of-domain queries without producing false positives.

The intuitive 3D visualization helps clinicians quickly understand symptom-organ associations and can serve as an effective educational tool in clinical settings.
In addition, it may assist in patient communication by providing visual explanations of symptom-related organ involvement.
Further potential applications and directions for future work are discussed in Sec.~\ref{sec:Conclusions and Future work}.

\section{Conclusions and Future Work}
\label{sec:Conclusions and Future work}

\begin{figure}[t]
    \centering
    \includegraphics[width=\linewidth]{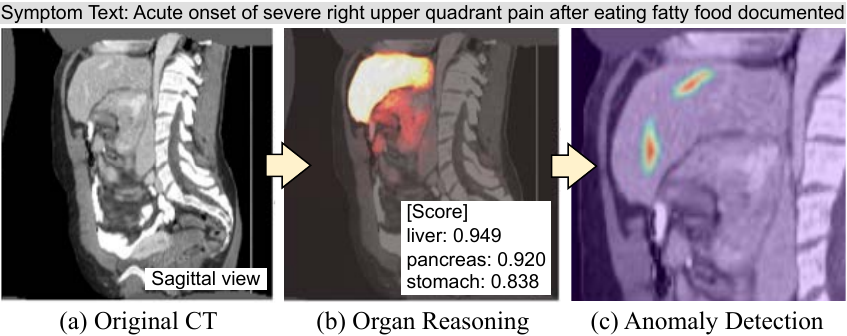}
    \caption{Future work directions for Med-SORA.}
    \label{fig:future_works}
\vspace{-15pt}
\end{figure}

In this paper, we proposed Med-SORA, a framework for symptom-to-organ reasoning that infers symptom-related organs in abdominal CT images from patient symptom text. 
We constructed a RAG-based dataset from medical literature and applied soft labeling with learnable organ anchors to model one-to-many symptom–organ relationships. 
To understand anatomical structures more effectively, we designed a 2D–3D cross-attention architecture that integrates both local and global CT features. 
Experiments show that Med-SORA outperforms existing multimodal medical models and enables interpretable 3D visualizations aligned with clinical reasoning.

This study has several limitations that suggest directions for future work. 
First, Med-SORA learns each organ independently, which limits holistic understanding of anatomical context. 
Second, the current framework provides only organ-level reasoning without capturing fine-grained organ details. 
Future extensions include joint multi-organ learning, detailed visualization, and automatic anomaly detection within related organ regions. 
These improvements could enhance clinical utility, and examples of such extensions are illustrated in Fig.~\ref{fig:future_works}.

{
    \small
    \bibliographystyle{ieeenat_fullname}
    \bibliography{main}

@String(AAAI = {AAAI})

@inproceedings{jin2019pubmedqa,
  title={PubMedQA: A Dataset for Biomedical Research Question Answering},
  author={Jin, Qiao and Dhingra, Bhuwan and Liu, Zhengping and Cohen, William and Lu, Xinghua},
  booktitle={Proceedings of the 2019 Conference on Empirical Methods in Natural Language Processing and the 9th International Joint Conference on Natural Language Processing (EMNLP-IJCNLP)},
  pages={2567--2577},
  year={2019}
}

@inproceedings{tran20153dcnn,
  title={Learning spatiotemporal features with 3d convolutional networks},
  author={Tran, Du and Bourdev, Lubomir and Fergus, Rob and Torresani, Lorenzo and Paluri, Manohar},
  booktitle={Proceedings of the IEEE international conference on computer vision},
  pages={4489--4497},
  year={2015}
}

@inproceedings{radford2021clip,
  title={Learning transferable visual models from natural language supervision},
  author={Radford, Alec and Kim, Jong Wook and Hallacy, Chris and Ramesh, Aditya and Goh, Gabriel and Agarwal, Sandhini and Sastry, Girish and Askell, Amanda and Mishkin, Pamela and Clark, Jack and others},
  booktitle={International conference on machine learning},
  pages={8748--8763},
  year={2021},
  organization={PmLR}
}

@inproceedings{devlin2019bert,
  title={Bert: Pre-training of deep bidirectional transformers for language understanding},
  author={Devlin, Jacob and Chang, Ming-Wei and Lee, Kenton and Toutanova, Kristina},
  booktitle={Proceedings of the 2019 conference of the North American chapter of the association for computational linguistics: human language technologies, volume 1 (long and short papers)},
  pages={4171--4186},
  year={2019}
}

@article{gu2021pubmedbert,
  title={Domain-specific language model pretraining for biomedical natural language processing},
  author={Gu, Yu and Tinn, Robert and Cheng, Hao and Lucas, Michael and Usuyama, Naoto and Liu, Xiaodong and Naumann, Tristan and Gao, Jianfeng and Poon, Hoifung},
  journal={ACM Transactions on Computing for Healthcare (HEALTH)},
  volume={3},
  number={1},
  pages={1--23},
  year={2021},
  publisher={ACM New York, NY}
}

@article{zhang2024biomedgpt,
  title={A generalist vision--language foundation model for diverse biomedical tasks},
  author={Zhang, Kai and Zhou, Rong and Adhikarla, Eashan and Yan, Zhiling and Liu, Yixin and Yu, Jun and Liu, Zhengliang and Chen, Xun and Davison, Brian D and Ren, Hui and others},
  journal={Nature Medicine},
  pages={1--13},
  year={2024},
  publisher={Nature Publishing Group US New York}
}

@inproceedings{eslami2023pubmedclip,
  title={Pubmedclip: How much does clip benefit visual question answering in the medical domain?},
  author={Eslami, Sedigheh and Meinel, Christoph and De Melo, Gerard},
  booktitle={Findings of the Association for Computational Linguistics: EACL 2023},
  pages={1181--1193},
  year={2023}
}

@article{zhang2023biomedclip,
  title={Biomedclip: a multimodal biomedical foundation model pretrained from fifteen million scientific image-text pairs},
  author={Zhang, Sheng and Xu, Yanbo and Usuyama, Naoto and Xu, Hanwen and Bagga, Jaspreet and Tinn, Robert and Preston, Sam and Rao, Rajesh and Wei, Mu and Valluri, Naveen and others},
  journal={arXiv preprint arXiv:2303.00915},
  year={2023}
}

@inproceedings{pal2022medmcqa,
  title={Medmcqa: A large-scale multi-subject multi-choice dataset for medical domain question answering},
  author={Pal, Ankit and Umapathi, Logesh Kumar and Sankarasubbu, Malaikannan},
  booktitle={Conference on health, inference, and learning},
  pages={248--260},
  year={2022},
  organization={PMLR}
}

@inproceedings{yu2024cp,
  title={Cp-clip: Core-periphery feature alignment clip for zero-shot medical image analysis},
  author={Yu, Xiaowei and Wu, Zihao and Zhang, Lu and Zhang, Jing and Lyu, Yanjun and Zhu, Dajiang},
  booktitle={International Conference on Medical Image Computing and Computer-Assisted Intervention},
  pages={88--97},
  year={2024},
  organization={Springer}
}

@inproceedings{hu2024omnimedvqa,
  title={Omnimedvqa: A new large-scale comprehensive evaluation benchmark for medical lvlm},
  author={Hu, Yutao and Li, Tianbin and Lu, Quanfeng and Shao, Wenqi and He, Junjun and Qiao, Yu and Luo, Ping},
  booktitle={Proceedings of the IEEE/CVF Conference on Computer Vision and Pattern Recognition},
  pages={22170--22183},
  year={2024}
}

@inproceedings{gao2024training,
  title={Training like a medical resident: Context-prior learning toward universal medical image segmentation},
  author={Gao, Yunhe},
  booktitle={Proceedings of the IEEE/CVF Conference on Computer Vision and Pattern Recognition},
  pages={11194--11204},
  year={2024}
}

@article{lewis2020retrieval,
  title={Retrieval-augmented generation for knowledge-intensive nlp tasks},
  author={Lewis, Patrick and Perez, Ethan and Piktus, Aleksandra and Petroni, Fabio and Karpukhin, Vladimir and Goyal, Naman and K{\"u}ttler, Heinrich and Lewis, Mike and Yih, Wen-tau and Rockt{\"a}schel, Tim and others},
  journal={Advances in neural information processing systems},
  volume={33},
  pages={9459--9474},
  year={2020}
}

@article{jin2023medcpt,
  title={Medcpt: Contrastive pre-trained transformers with large-scale pubmed search logs for zero-shot biomedical information retrieval},
  author={Jin, Qiao and Kim, Won and Chen, Qingyu and Comeau, Donald C and Yeganova, Lana and Wilbur, W John and Lu, Zhiyong},
  journal={Bioinformatics},
  volume={39},
  number={11},
  pages={btad651},
  year={2023},
  publisher={Oxford University Press}
}

@article{liu2021auto,
  title={Auto-encoding knowledge graph for unsupervised medical report generation},
  author={Liu, Fenglin and You, Chenyu and Wu, Xian and Ge, Shen and Sun, Xu and others},
  journal={Advances in Neural Information Processing Systems},
  volume={34},
  pages={16266--16279},
  year={2021}
}

@article{zakka2024almanac,
  title={Almanac—retrieval-augmented language models for clinical medicine},
  author={Zakka, Cyril and Shad, Rohan and Chaurasia, Akash and Dalal, Alex R and Kim, Jennifer L and Moor, Michael and Fong, Robyn and Phillips, Curran and Alexander, Kevin and Ashley, Euan and others},
  journal={Nejm ai},
  volume={1},
  number={2},
  pages={AIoa2300068},
  year={2024},
  publisher={Massachusetts Medical Society}
}

@inproceedings{huy2025interactive,
  title={Interactive Medical Image Analysis with Concept-based Similarity Reasoning},
  author={Huy, Ta Duc and Tran, Sen Kim and Nguyen, Phan and Tran, Nguyen Hoang and Sam, Tran Bao and van den Hengel, Anton and Liao, Zhibin and Verjans, Johan W and To, Minh-Son and Phan, Vu Minh Hieu},
  booktitle={Proceedings of the Computer Vision and Pattern Recognition Conference},
  pages={30797--30806},
  year={2025}
}

@inproceedings{ko2025bringing,
  title={Bringing CLIP to the Clinic: Dynamic Soft Labels and Negation-Aware Learning for Medical Analysis},
  author={Ko, Hanbin and Park, Chang-Min},
  booktitle={Proceedings of the Computer Vision and Pattern Recognition Conference},
  pages={25897--25906},
  year={2025}
}

@article{oord2018infonce,
  title={Representation learning with contrastive predictive coding},
  author={Oord, Aaron van den and Li, Yazhe and Vinyals, Oriol},
  journal={arXiv preprint arXiv:1807.03748},
  year={2018}
}

@inproceedings{gao2024softclip,
  title={Softclip: Softer cross-modal alignment makes clip stronger},
  author={Gao, Yuting and Liu, Jinfeng and Xu, Zihan and Wu, Tong and Zhang, Enwei and Li, Ke and Yang, Jie and Liu, Wei and Sun, Xing},
  booktitle={Proceedings of the AAAI Conference on Artificial Intelligence},
  volume={38},
  number={3},
  pages={1860--1868},
  year={2024}
}

@article{huang2019clinicalbert,
  title={Clinicalbert: Modeling clinical notes and predicting hospital readmission},
  author={Huang, Kexin and Altosaar, Jaan and Ranganath, Rajesh},
  journal={arXiv preprint arXiv:1904.05342},
  year={2019}
}

@article{pubmed,
  title={https://pubmed.ncbi.nlm.nih.gov/},
  year={1996}
}

@article{wikipedia,
  title={https://www.wikipedia.org/},
  year={2001}
}

@inproceedings{wu2025graph,
  title={Medical Graph RAG: Evidence-based Medical Large Language Model via Graph Retrieval-Augmented Generation},
  author={Wu, Junde and Zhu, Jiayuan and Qi, Yunli and Chen, Jingkun and Xu, Min and Menolascina, Filippo and Jin, Yueming and Grau, Vicente},
  booktitle={Proceedings of the 63rd Annual Meeting of the Association for Computational Linguistics (Volume 1: Long Papers)},
  pages={28443--28467},
  year={2025}
}

@article{xu2025structure,
  title={Structure Causal Models and LLMs Integration in Medical Visual Question Answering},
  author={Xu, Zibo and Li, Qiang and Nie, Weizhi and Wang, Weijie and Liu, Anan},
  journal={IEEE Transactions on Medical Imaging},
  year={2025},
  publisher={IEEE}
}

@article{nori2025sequential,
  title={Sequential Diagnosis with Language Models},
  author={Nori, Harsha and Daswani, Mayank and Kelly, Christopher and Lundberg, Scott and Ribeiro, Marco Tulio and Wilson, Marc and Liu, Xiaoxuan and Sounderajah, Viknesh and Carlson, Jonathan and Lungren, Matthew P and others},
  journal={arXiv preprint arXiv:2506.22405},
  year={2025}
}

@article{shaw2025artificial,
  title={Artificial intelligence in medical education: a scoping review of the evidence for efficacy and future directions},
  author={Shaw, Kody and Henning, Marcus A and Webster, Craig S},
  journal={Medical Science Educator},
  pages={1--14},
  year={2025},
  publisher={Springer}
}

@inproceedings{btcv2015miccai,
  title={Miccai multi-atlas labeling beyond the cranial vault--workshop and challenge},
  author={Landman, Bennett and Xu, Zhoubing and Igelsias, Juan and Styner, Martin and Langerak, Thomas and Klein, Arno},
  booktitle={Proc. MICCAI multi-atlas labeling beyond cranial vault—workshop challenge},
  volume={5},
  pages={12},
  year={2015},
  organization={Munich, Germany}
}

@article{ridnik2021imagenet,
  title={Imagenet-21k pretraining for the masses},
  author={Ridnik, Tal and Ben-Baruch, Emanuel and Noy, Asaf and Zelnik-Manor, Lihi},
  journal={arXiv preprint arXiv:2104.10972},
  year={2021}
}

@inproceedings{hoffer2015deep,
  title={Deep metric learning using triplet network},
  author={Hoffer, Elad and Ailon, Nir},
  booktitle={Similarity-based pattern recognition: third international workshop, SIMBAD 2015, Copenhagen, Denmark, October 12-14, 2015. Proceedings 3},
  pages={84--92},
  year={2015},
  organization={Springer}
}

@article{neelakantan2022openai,
  title={Text and code embeddings by contrastive pre-training},
  author={Neelakantan, Arvind and Xu, Tao and Puri, Raul and Radford, Alec and Han, Jesse Michael and Tworek, Jerry and Yuan, Qiming and Tezak, Nikolas and Kim, Jong Wook and Hallacy, Chris and others},
  journal={arXiv preprint arXiv:2201.10005},
  year={2022}
}

@article{krithara2023bioasq,
  title={BioASQ-QA: A manually curated corpus for Biomedical Question Answering},
  author={Krithara, Anastasia and Nentidis, Anastasios and Bougiatiotis, Konstantinos and Paliouras, Georgios},
  journal={Scientific Data},
  volume={10},
  number={1},
  pages={170},
  year={2023},
  publisher={Nature Publishing Group UK London}
}

@article{radford2018GPT,
  title={Improving language understanding by generative pre-training},
  author={Radford, Alec and Narasimhan, Karthik and Salimans, Tim and Sutskever, Ilya and others},
  year={2018},
  publisher={San Francisco, CA, USA}
}

@inproceedings{dosovitskiy2020vit,
  title={An Image is Worth 16x16 Words: Transformers for Image Recognition at Scale},
  author={Dosovitskiy, Alexey and Beyer, Lucas and Kolesnikov, Alexander and Weissenborn, Dirk and Zhai, Xiaohua and Unterthiner, Thomas and Dehghani, Mostafa and Minderer, Matthias and Heigold, G and Gelly, S and others},
  booktitle={International Conference on Learning Representations},
  year={2020}
}

@article{singhal2025medpalm,
  title={Toward expert-level medical question answering with large language models},
  author={Singhal, Karan and Tu, Tao and Gottweis, Juraj and Sayres, Rory and Wulczyn, Ellery and Amin, Mohamed and Hou, Le and Clark, Kevin and Pfohl, Stephen R and Cole-Lewis, Heather and others},
  journal={Nature Medicine},
  pages={1--8},
  year={2025},
  publisher={Nature Publishing Group US New York}
}
}


\end{document}